\documentclass[letterpaper]{article} 
\usepackage{aaai2026}  
\usepackage{times}  
\usepackage{helvet}  
\usepackage{courier}  
\usepackage[hyphens]{url}  
\usepackage{graphicx} 
\usepackage{natbib}  
\usepackage{caption} 
\usepackage{algorithm}
\usepackage{algorithmic}
\usepackage{tabularx}

\usepackage{amsmath}
\usepackage{array}
\usepackage[caption=false,font=normalsize,labelfont=sf,textfont=sf]{subfig}
\usepackage{textcomp}
\usepackage{verbatim}
\usepackage{cite}
\usepackage{amssymb}
\usepackage{booktabs}
\usepackage{enumitem}
\usepackage{epsfig}
\usepackage{amsfonts}
\usepackage{bm}
\usepackage{multirow}
\usepackage{makecell}
\usepackage{bbding}
\usepackage[table]{xcolor}
\usepackage{tikz}
\usepackage{newfloat}
\usepackage{listings}
\DeclareCaptionStyle{ruled}{labelfont=normalfont,labelsep=colon,strut=off} 
\floatstyle{ruled}
\newfloat{listing}{tb}{lst}{}
\floatname{listing}{Listing}
\title{PulseMind: A Multi-Modal Medical Model for Real-World Clinical Diagnosis}
\author{
    Jiao Xu\textsuperscript{\rm 1,2}\thanks{Equal contribution. Work performed when Jiao Xu was an intern of Ant Group.},
    Junwei Liu\textsuperscript{\rm 2,3}\footnotemark[1],
    Jiangwei Lao\textsuperscript{\rm 2},
    Qi Zhu\textsuperscript{\rm 2},
    Yunpeng Zhao\textsuperscript{\rm 4},
    Congyun Jin\textsuperscript{\rm 2},
    Shinan Liu\textsuperscript{\rm 4},
    Zhihong Lu\textsuperscript{\rm 2},
    Lihe Zhang\textsuperscript{\rm 1}\thanks{Corresponding authors.},
    Xin Chen\textsuperscript{\rm 5}\footnotemark[2],
    Jian Wang\textsuperscript{\rm 2}\footnotemark[2],
    Ping Wang\textsuperscript{\rm 3}\footnotemark[2]
}
\affiliations {
    \textsuperscript{\rm 1}	Dalian University of Technology\\
    \textsuperscript{\rm 2} Ant Group\\
    \textsuperscript{\rm 3} Peking University\\
    \textsuperscript{\rm 4} University of Hong Kong\\
    \textsuperscript{\rm 5} City University of Hong Kong\\
    xjmmcome@mail.dlut.edu.cn,
    wenshuo.ljw@antgroup.com,
    zhanglihe@dlut.edu.cn
}

\begin{document}

\maketitle

\begin{abstract}

Recent advances in medical multi-modal models focus on specialized image analysis like dermatology, pathology, or radiology. However, they do not fully capture the complexity of real-world clinical diagnostics, which involve heterogeneous inputs and require ongoing contextual understanding during patient-physician interactions.
To bridge this gap, we introduce PulseMind, a new family of multi-modal diagnostic models that integrates a systematically curated dataset, a comprehensive evaluation benchmark, and a tailored training framework.
Specifically, we first construct a diagnostic dataset, MediScope, which comprises 98,000 real-world multi-turn consultations and 601,500 medical images, spanning over 10 major clinical departments and more than 200 sub-specialties.
Then, to better reflect the requirements of real-world clinical diagnosis, we develop the PulseMind Benchmark, a multi-turn diagnostic consultation benchmark with a four-dimensional evaluation protocol comprising proactiveness, accuracy, usefulness, and language quality.
Finally, we design a training framework tailored for multi-modal clinical diagnostics, centered around a core component named Comparison-based Reinforcement Policy Optimization (CRPO). Compared to absolute score rewards, CRPO uses relative preference signals from multi-dimensional comparisons to provide stable and human-aligned training guidance.
Extensive experiments demonstrate that PulseMind achieves competitive performance on both the diagnostic consultation benchmark and public medical benchmarks.
\end{abstract} 

\begin{links}
\link{Code}{https://github.com/AQ-MedAI/PulseMind}
\end{links}

\section{Introduction}

\begin{figure}[t]
\begin{center}
\includegraphics[width=1\linewidth]{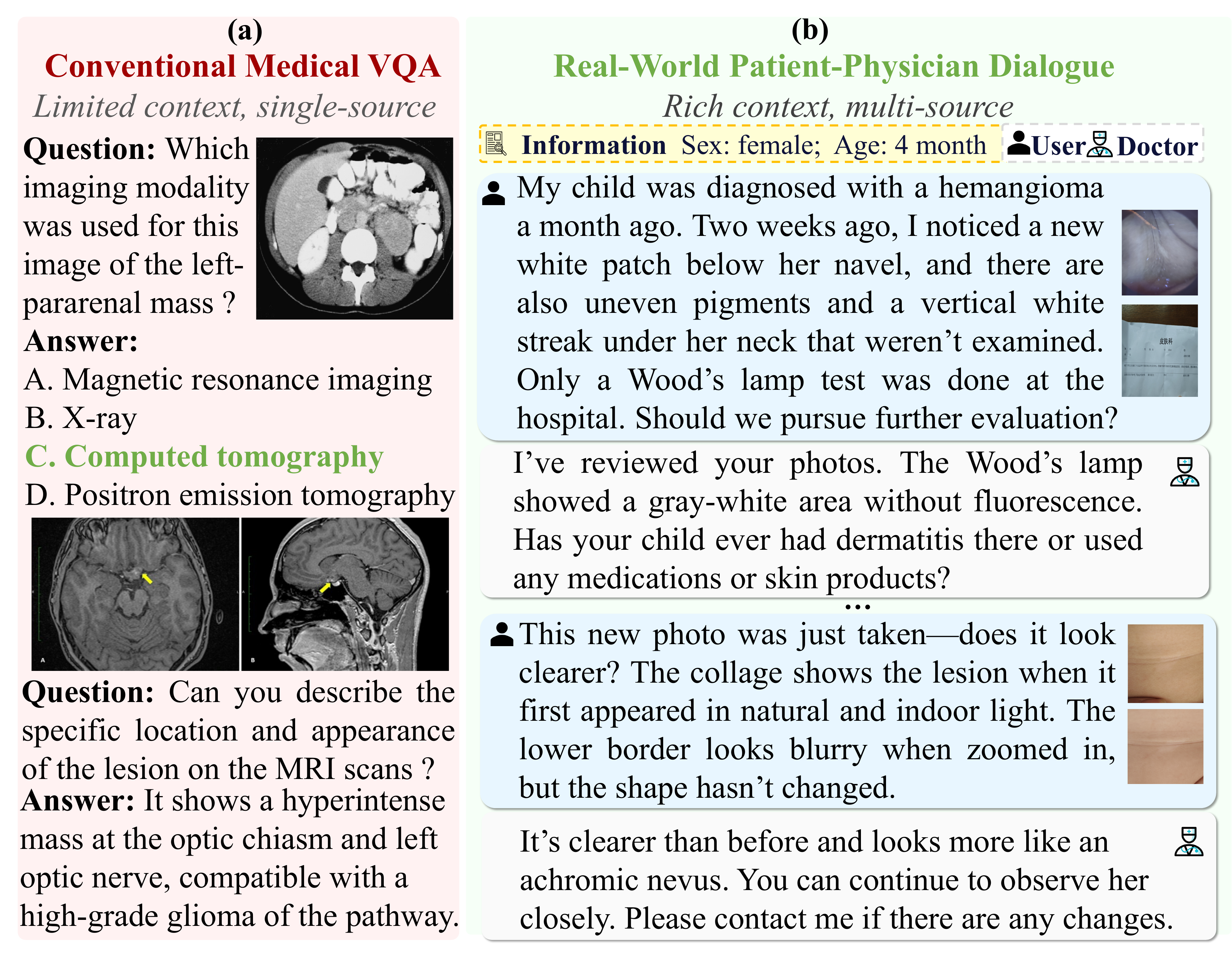}
\end{center}
   \caption{ Comparison between (a) conventional multi-modal medical VQA tasks and (b) real-world diagnostic dialogue scenarios.}
\label{fig:show}
\end{figure}

In recent years, Visual Language Models (VLMs) \cite{gpt4,llava,qwenvl,internvl3,gemini} have made remarkable progress across various fields due to their strong capabilities in visual understanding and multi-modal reasoning. These advances have inspired the development of medical multi-modal large models~\cite{llava-med,med-gemini,medvlm,medgemma,xraygpt,Med-flamingo}. However, despite this progress, several critical challenges remain insufficiently addressed, especially in real-world clinical applications.

Compared to conventional medical image analysis (Fig.\ref{fig:show}(a)), real-world diagnostic consultations (Fig.\ref{fig:show}(b)) present fundamentally different and significantly complex characteristics. These real-world scenarios involve not only the integration of heterogeneous information from multiple imaging modalities, but also the handling of multi-turn interactions between physicians and patients.
This complexity exposes two fundamental gaps in current research:

\textbf{i)} \textit{Limitations in training datasets.} Most existing datasets either focus on visual question answering (VQA)~\cite{pathvqa,VQA-RAD,huatuo-vision} or contain only a single image modality, lacking both the diversity of visual inputs and the multi-turn dialogue context that are critical in clinical consultations.

\textbf{ii)} \textit{Inadequate evaluation benchmarks.} 
As a recent study~\cite{lingshu} pointed out, current medical multi-modal benchmarks fall short of capturing real-world clinical complexity, limiting their ability to evaluate model utility in practical downstream tasks, such as clinical diagnostics. Although a few benchmarks include dialogue cases~\cite{healthbench}, these are mostly limited to pure text and lack integration with medical imagery.

To address these challenges, we propose PulseMind, a new multi-modal diagnostic model that encompasses a systematically curated dataset, an evaluation benchmark, and a tailored training framework.

For the dataset construction, we systematically curated a large-scale multi-modal diagnostic dialogue dataset through the following four key steps: collection, anonymization, expansion, and proofreading.
The final dataset consists of 98,000 real-world multi-turn consultation dialogues and 601,500 medical images, spanning over 10 major clinical departments and 200 sub-specialties. It also incorporates a broad spectrum of clinical data types, including laboratory test results, examination reports, prescriptions, medical images, surgical records, and other relevant reports, demonstrating strong clinical diversity and representativeness.
For the diagnostic dialogue evaluation, we introduce the PulseMind Benchmark, which features multi-turn scenarios incorporating both plain-text and multi-modal inputs, enabling a comprehensive assessment of diagnostic dialogue capabilities. To better simulate real-world clinical settings, we further develop a GPT-based automatic evaluation framework that assesses model performance across four key dimensions: accuracy, proactiveness, usefulness, and language quality, providing a well-rounded metric for evaluating the quality of diagnostic interactions.

During the model training phase, we first perform supervised fine-tuning (SFT) on medical text and multi-modal data to build domain knowledge and enhance multi-modal understanding. Following this, to better optimize diagnostic responses, we employ a reinforcement learning stage using our proposed Comparison-based Reinforcement Policy Optimization (CRPO) method. 
The motivation behind CRPO stems from our observation that using absolute numerical scores as rewards often leads to instability and does not align well with human preferences. Humans find it easier to judge which of two responses is better rather than assigning an absolute quality score to a single response. To leverage this insight, CRPO trains the model by comparing its responses pairwise against those from multiple counterpart models. These comparisons evaluate key aspects including proactiveness, accuracy, usefulness, and language quality. By learning from relative preference signals instead of absolute rewards, CRPO provides more stable and human-aligned guidance, ultimately encouraging the model to generate more reliable and higher-quality diagnostic responses.

Experimental results show that PulseMind achieves an average win rate of 76\% on the PulseMind Benchmark, indicating its promising diagnostic consultation capabilities. Furthermore, the model achieves competitive performance on 11 public medical question-answering datasets, demonstrating solid generalization capabilities.

In summary, the contributions of this work are three-fold:
\begin{itemize}[leftmargin=0.468cm]
\item We introduce the \textit{MediScope Dataset}, the first large-scale multi-modal clinical diagnostics dataset featuring rich, real-world multi-turn diagnostic consultations.
\item We propose the \textit{PulseMind Benchmark}, the first benchmark for evaluating clinical diagnostic capabilities, providing a comprehensive multi-dimensional assessment of model in multi-turn diagnostic consultations.
\item We present \textit{PulseMind}, a medical multi-modal large model specifically designed for real-world clinical diagnostics, achieving competitive performance on both the PulseMind Benchmark and public medical question-answering datasets.
\end{itemize}

\section{Related Work}

\subsection{General Visual Language Models}
Visual Language Models (VLMs) have emerged as a new paradigm for general-purpose artificial intelligence, demonstrating impressive capabilities in  multi-modal understanding by leveraging large-scale datasets. The InternVL series \cite{internvl, internvl3} advances multi-modal learning through the expansion of visual encoders, thereby improving the representation of fine-grained visual details. The Qwen-VL models \cite{qwen2, qwenvl} build upon strong large language models, showing notable strengths in contextual understanding and instruction following. The OpenAI “o” series \cite{o1} and DeepSeek-R1 \cite{deepseek-v3, deepseek-r1} further push performance boundaries via post-training alignment techniques. Meanwhile, closed-source flagship models such as GPT-4o \cite{gpt-4o}, Gemini 2.5 Pro \cite{gemini}, and Claude 3.5 \cite{claude-3.5} exhibit strong performance in cross-modal reasoning and dialogue generation, and are often used as reference points for evaluating the capabilities of current VLM systems.

However, despite their success in open-domain tasks, existing models are not readily transferable to the medical domain due to the specialized nature of clinical information \cite{danger1,danger2}. In the absence of sufficient medical knowledge, these models often misinterpret critical diagnostic features, resulting in potentially inaccurate or unreliable predictions \cite{hallucination1,hallucination2}, which may pose risks to patient safety. These challenges underscore the necessity of developing dedicated multi-modal large models tailored to medical applications.

\begin{figure*}[t]
\begin{center}
\includegraphics[width=0.95\linewidth]{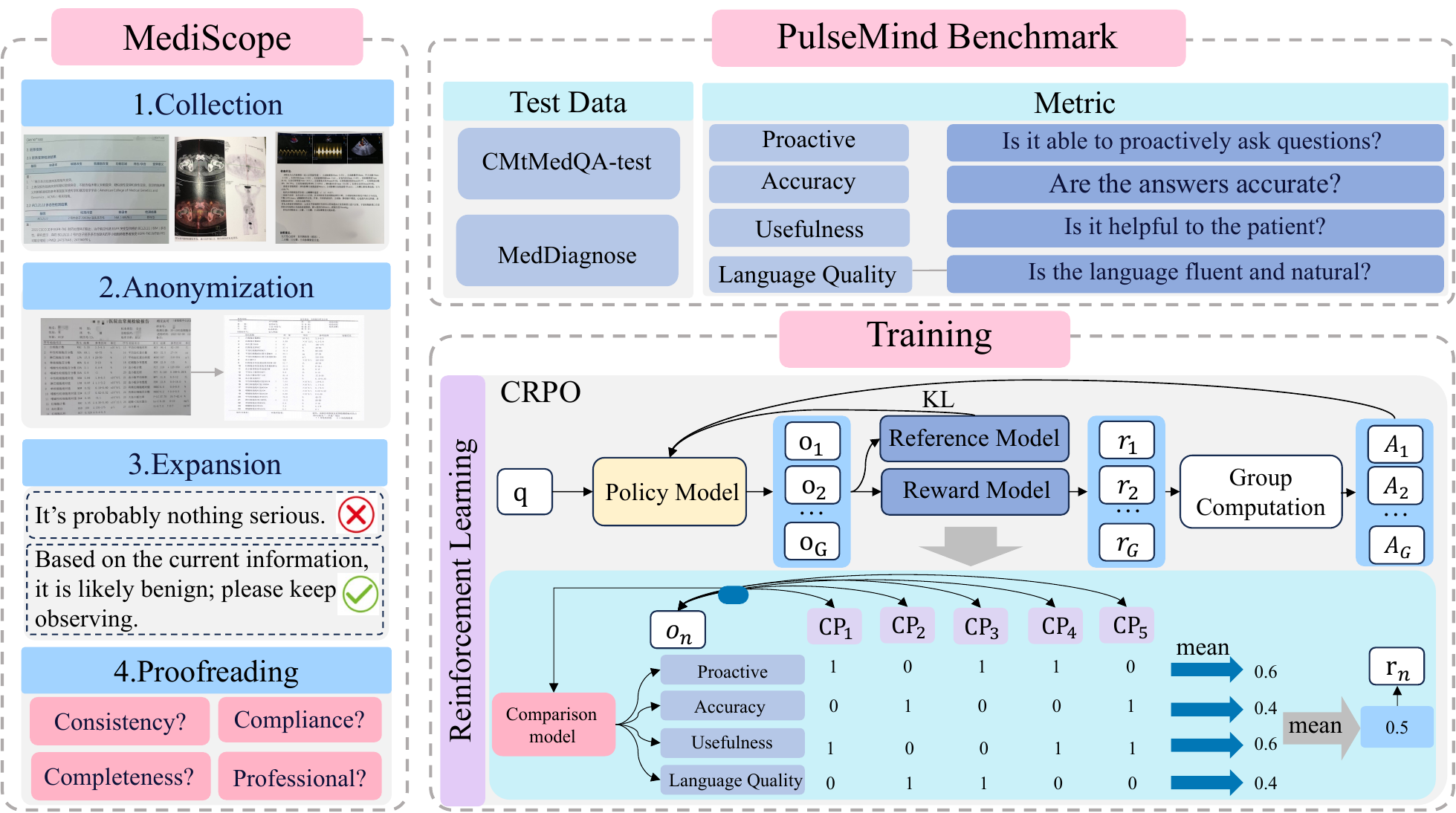}
\end{center}
   \caption{Overview of PulseMind, including dataset construction (MediScope), the PulseMind Benchmark, and CRPO training.``CP" denotes the counterpart model.} 
\label{fig:framework}
\end{figure*}

\subsection{Medical Visual Language Models}

Recent years have witnessed significant advances in medical visual language models, aiming to adapt general-purpose architectures for clinical applications. Early efforts such as LLaVA-Med \cite{llava-med} and Med-VLM \cite{medvlm} enhanced general models by incorporating medical knowledge via large-scale fine-tuning. Subsequently, domain-specific models have been developed for specialized fields including radiology \cite{xraygpt, chest-xray}, pathology \cite{slidechat,patho-r1}, surgery \cite{surgvlm}, ophthalmology \cite{eyeclip}, and dermatology \cite{PanDerm}, achieving notable accuracy in their respective domains. 
More recently, general-purpose medical VLMs such as HuatuoGPT-Vision \cite{huatuo-vision}, Med-GEMMA \cite{medgemma}, Lingshu \cite{lingshu}, and Med-Gemini \cite{med-gemini} have been developed, trained on large-scale multi-specialty datasets to strike a balance between broad generalization and specialized domain expertise.

Despite these advances, most existing models primarily focus on specialized image analysis and fall short of addressing the complexity inherent in real-world clinical diagnostics. Such diagnostics require the integration of heterogeneous multi-modal inputs and the maintenance of contextual coherence over multi-turn patient-physician interactions. To tackle these challenges, we propose PulseMind, a new multi-modal diagnostic model specifically designed for realistic clinical dialogue scenarios.

\section{PulseMind}

We propose PulseMind, a unified framework tailored for multi-modal physician-patient consultation scenarios. As shown in Fig.~\ref{fig:framework}, our framework consists of three core components. First, we construct a large-scale multi-modal medical dialogue dataset to provide a solid data foundation for model training. Second, we design a comprehensive evaluation benchmark that realistically reflects the key challenges in clinical consultations. Finally, we introduce a Comparison-based Reinforcement Policy Optimization (CRPO) method to conduct reinforcement learning training.

\subsection{Dataset} 

\begin{figure*}[t]
\begin{center}
\includegraphics[width=0.95\linewidth]{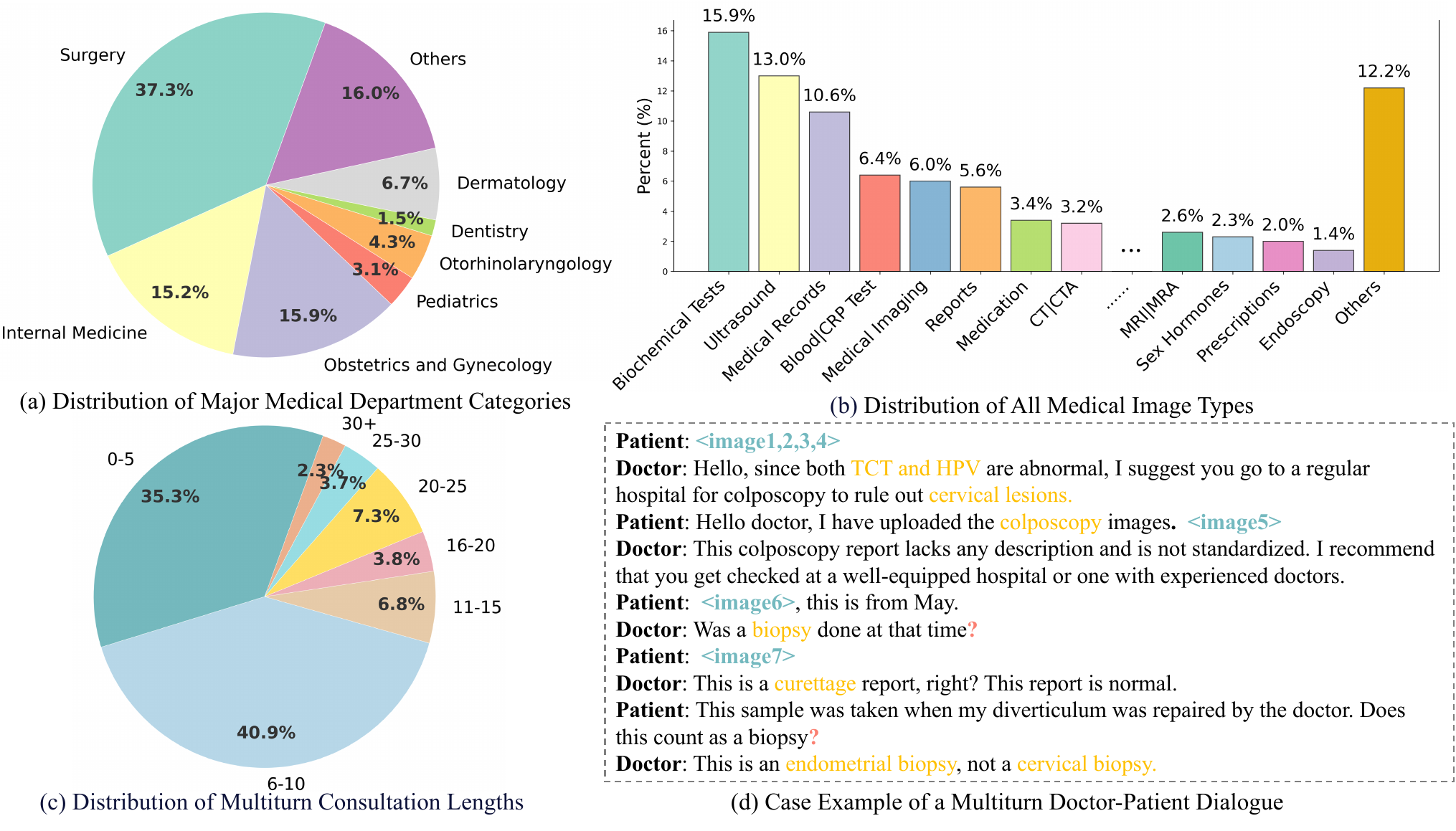}
\end{center}
   \caption{ Characteristics of the collected multi-turn dialogue training set from four perspectives: (a) Distribution of major departments; (b) Heterogeneous Medical Image Modalities; (c) Distribution of multi-turn dialogue lengths; (d) Example of a multi-turn physician–patient consultation dialogue.} 
\label{fig:data}
\end{figure*}

Our training data consists of our self-constructed dataset, MediScope, along with a collection of publicly available textual and multi-modal datasets, totaling approximately 792,000 samples.
\subsubsection{MediScope.} We construct a large-scale, heterogeneous multi-modal dataset that captures the complexity of clinical dialogues, referred to as MediScope.
The construction process follows a rigorous four-stage pipeline:
i) Collection: We collect de-identified data from real-world clinical scenarios, encompassing diverse modalities such as examination reports and multi-turn physician-patient dialogues that authentically capture the complexity and variability of diagnostic processes.
ii) Anonymization: We apply advanced Optical Character Recognition (OCR) and Named Entity Recognition (NER) techniques to perform a secondary anonymization check on both text and images, ensuring complete removal of personally identifiable information and compliance with privacy regulations.
iii) Expansion: We use large language models~\cite{gpt-4o,gemini} to refine and expand physician responses by filtering out meaningless fillers and augmenting clinically relevant content, enhancing the dialogue’s completeness, clarity, and coherence without compromising medical accuracy or intent.
iv) Proofreading: Medical experts and licensed physicians thoroughly review and refine the expanded dialogues to ensure clinical validity, ethical compliance, and appropriate expression of empathy. As a result, the dataset comprises 98,000 real-world multi-turn consultations and 601,500 medical images, spanning over 10 major clinical departments and more than 200
sub-specialties. Moreover, it includes a wide variety of data
types such as laboratory test results, examination reports,
prescriptions, medical images, and surgical records, reflecting strong clinical diversity and representativeness.

Fig.~\ref{fig:data} provides an overview of the dataset's key characteristics across multiple dimensions.
Specifically, Fig.~\ref{fig:data}(a) shows the distribution across a wide range of clinical departments, highlighting the dataset’s broad coverage.
Fig.~\ref{fig:data}(b) illustrates the diversity of medical image types, including Ultrasound, Medical Records, CT/MRI, Pathology, and Endoscopy, which supports comprehensive vision-language learning across multiple modalities.
Fig.~\ref{fig:data}(c) shows that the dataset contains a large proportion of multi-turn dialogues, with 40.9\% of dialogues containing 6–10 turns and 6\% exceeding 20 turns, facilitating long-range dependency modeling and contextual reasoning.
Finally, Fig.~\ref{fig:data}(d) presents a real-world physician-patient conversation that exemplifies the nature of clinical interactions.

\subsubsection{Public Datasets.} Complementing MediScope, we incorporate a diverse collection of publicly available medical datasets to strengthen the model’s capability in traditional medical image analysis. Specifically, we leverage both text-only datasets \cite{huatuo26m,PubMedVision,medQA,medmcqa,medxpertqa,mmlu,CMtMedQA} and multi-modal resources \cite{dermavqa,pathvqa,pmc-VQA,slake,VQA-RAD,mmmu,medxpertqa}.

\begin{figure*}[t]
\begin{center}
\includegraphics[width=0.98\linewidth]{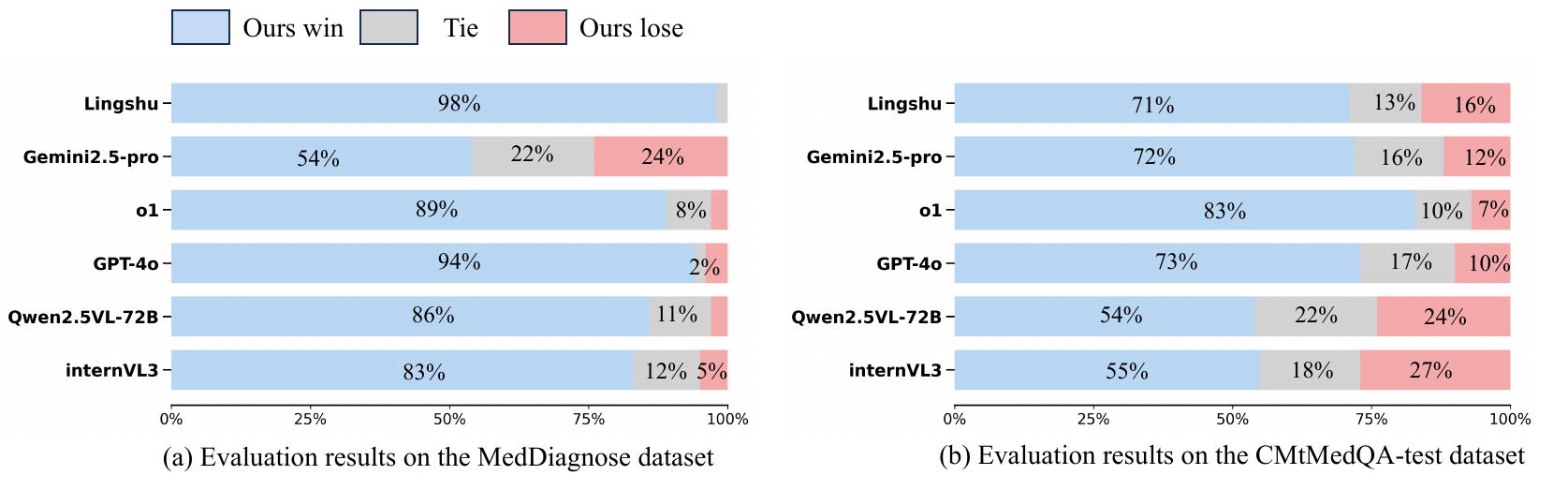}
\end{center}
   \caption{Win rates of our model against six baseline methods on the PulseMind Benchmark} 
\label{fig:diagnose}
\end{figure*}

\begin{figure*}[t]
\begin{center}
\includegraphics[width=0.95\linewidth]{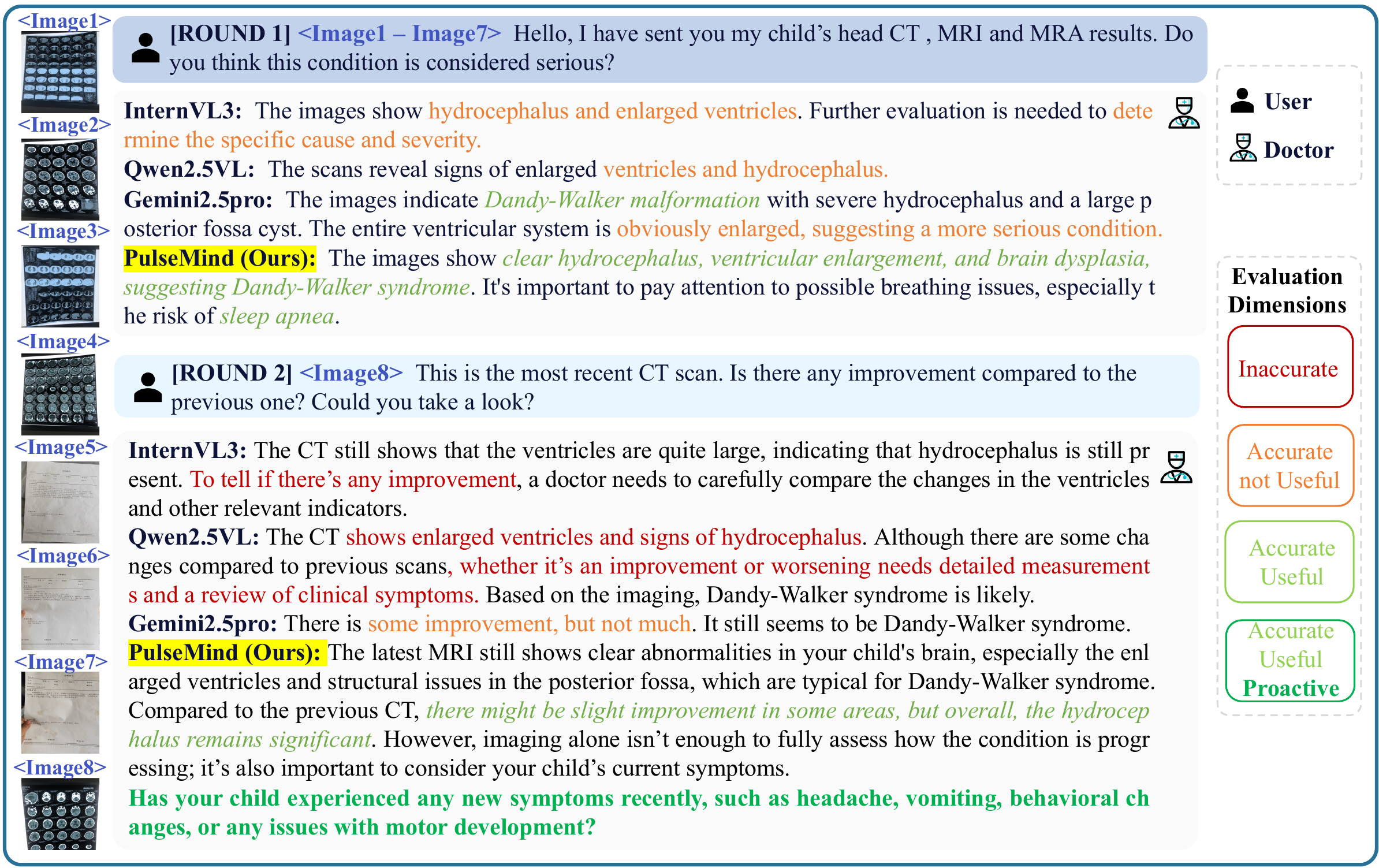}
\end{center}
   \caption{Illustrative cases of six models on the PulseMind Benchmark. Representative response quality is color-coded, as indicated by the tags on the right.}
\label{fig:example}
\end{figure*}

\begin{table*}[t]\Huge
\resizebox{1\linewidth}{!}{
  \setlength{\tabcolsep}{1mm}{  
  \small
  \begin{tabular}{l| ccc ccc ccc ccc}
    \toprule
    
    \multirow{2}*{Method} &\multicolumn{7}{c}{Muilti-modal QA} & &\multicolumn{4}{c}{Text-only QA}\\
    \cline{2-8}
    \cline{10-13}
    &MMMU &VQA-RAD & PMC-VQA&  SLAKE &PathVQA &DermaVQA &MedXQA-MM &  &MMLU &MedMCQA &MedQA &MedXQA-text \\
    
    \bottomrule
     
    \multicolumn{13}{c}{\textit{Proprietary  Models}}\\
    \bottomrule
    GPT-4o	& 57.3&71.2	&55.2	&67.4 &55.5 &35.0 &22.3 &  & 88.7	&73.5 &55.7 &22.5 		 \\
    o1 &57.8  &63.0   &54.5   &69.9  &57.3 &43.0 &49.7 &  & 91.6	&82.7 & 86.6 &48.9  \\
    Gemini2.5-pro &49.3	&70.5	&55.5	&75.8 &55.4 &39.0   &39.5	&  &89.8 	&68.6 &85.6 &24.3   \\ 
    \bottomrule
    \multicolumn{13}{c}{\textit{Open-source Models ($\sim$72B)}}\\
    \bottomrule
    InternVL3-78B	& \underline{69.1}&73.6	&56.6	&77.4 &\underline{51.0} &\underline{37.0} &27.4 &  &83.0 	&66.1 &\underline{93.3} & \underline{18.5}		 \\
    Qwen2.5VL-72B  &66.4 &\underline{80.3}   &\underline{59.3}  &\underline{78.3}   &42.3 &34.0 &\underline{27.6} &  &\underline{88.3} 	&\underline{67.2} &91.3 &16.1 \\
    \rowcolor{gray!15}
     PulseMind-72B	&\textbf{69.4} &\textbf{87.1}	&\textbf{70.3}	&\textbf{85.6} & \textbf{64.9} &\textbf{42.0} &\textbf{36.7}&  &\textbf{88.7 }	&\textbf{71.3} &\textbf{94.8} &\textbf{29.8} 		  \\
    \bottomrule
    \multicolumn{13}{c}{\textit{Open-source Models ($\sim$32B)}}\\
    \bottomrule
    InternVL3-38B	&\textbf{65.2} &65.4	&56.6	&72.7 &51.0 &\underline{31.0}  &25.2 & &82.8 	&64.9 &73.5 &16.0 		 \\
    Qwen2.5VL-32B  &62.8 &73.8   &54.5  &71.2   &41.9 &25.0  &25.2 &  &83.2 	&63.0 &71.6 &15.6 \\
    LLAVA-med-34B	&48.9 &58.6	&44.4	&67.3 &48.8 &13.0 &16.4  & &74.7 	&52.2 &63.5 &14.1 	 \\   
    HuatuoGPT-vision-34B &54.3 &61.4  &56.6  &69.5   &44.4 &21.0 &17.3 &  &80.8 	&63.6 &57.4 &16.0 \\
    Lingshu-32B	&62.3 &\underline{76.5}	&\underline{57.9}	&\textbf{89.2}  &\textbf{65.9} &17.0  &\textbf{30.9} & &\underline{84.7} 	&\underline{66.1} &\underline{74.7} &\textbf{22.7} 	\\
    \rowcolor{gray!15}
     PulseMind-32B	& \underline{64.6} &\textbf{83.2} &\textbf{68.1}	&\underline{81.5} &\underline{62.0} & \textbf{32.0} & \underline{29.6}& &\textbf{85.6} 	& \textbf{66.4}&\textbf{92.9} &\underline{21.5} 		  \\

  \bottomrule
\end{tabular}
}}
\caption{Performance comparison on medical QA benchmarks. The top two results highlighted with \textbf{bold} and \underline{underlined} fonts, respectively. Rows with \colorbox{gray!15}{gray} background indicate our PulseMind models. ``MMMU" refers to MMMU Health \& Medicine, ``MedXQA" refers to MedXpertQA, and ``MMLU" refers to MMLU clinical topics.}
\label{tab-multi}
\end{table*}

\subsection{Evaluation Benchmark}

Recent studies~\cite{lingshu} have pointed out that current medical multi-modal benchmarks do not adequately capture the complexity of real-world clinical scenarios, which limits their ability to evaluate model performance in practical diagnostic tasks. To address this, we introduce the PulseMind Benchmark, which integrates multi-source images and multi-turn dialogue data to realistically simulate clinical workflows and assess diagnostic dialogue capabilities. Furthermore, to situate our results within the broader field and enable fair comparison with existing methods, we also evaluate our models on standard medical QA benchmarks.

\subsubsection{PulseMind Benchmark}

To comprehensively evaluate model capabilities in real-world medical consultation scenarios, we propose the PulseMind Benchmark, which integrates three core components:  
(i) a multi-source dataset covering both text-based and multi-modal multi-turn diagnostic dialogues;  
(ii) clinically driven multi-dimensional evaluation metrics;  
and (iii) a human-aligned relative scoring strategy.  
Together, these components establish a unified framework to systematically and rigorously assess model performance in authentic clinical environments.

\textit{Benchmark Composition.}  
The PulseMind Benchmark combines two datasets, covering both text-only and multi-modal consultation scenarios with over 1,200 samples in total. The multi-modal consultation set, named MedDiagnose, was constructed by us and contains 237 samples collected from patient cases, featuring images alongside expert-verified dialogues. For text-based dialogue understanding, we expand the original CMtMedQA-test dataset to include multi-turn reasoning, resulting in 1,000 samples. 

\textit{Multi-dimensional Evaluation Protocol.}  
Evaluating physician-patient consultation quality requires more than correctness, involving clinical reasoning, contextual interaction, and communicative clarity ~\cite{zhongjing,zhibiao}. To capture these aspects, we design an evaluation protocol based on four key dimensions:

\begin{itemize}
\item Proactiveness, which assesses whether the model actively inquires about missing but critical information, emulating the diagnostic behavior of experienced physicians.

\item Accuracy, which verifies whether the diagnostic suggestions are medically sound and free from factual errors or inappropriate reasoning.

\item Usefulness, which measures the practical value of the response, including clarity, actionability, and relevance to the patient’s concerns.

\item Language Quality, which evaluates fluency, professionalism, and overall communicative effectiveness.
\end{itemize}

These four dimensions jointly capture both clinical content quality and the effectiveness of medical communication, providing a holistic view of model performance.

\textit{Evaluation Strategy.}  
We adopt GPT-4 as an automatic evaluator, following recent studies~\cite{gptjudging1, gptjudging2}. For each input prompt, model responses are compared against those of multiple baselines across the four evaluation dimensions mentioned above. The outcomes are categorized as win, tie, or loss, and the win rate is used as the primary evaluation metric.

\subsubsection{General Medical QA Benchmarks.}
In addition to consultation evaluation, we also assess model performance on traditional medical question answering tasks in both multi-modal and text-only settings, using 11 benchmark datasets.

\textit{Benchmark Composition.}
For multi-modal QA, we select datasets that cover general medical knowledge~\cite{mmmu,pmc-VQA}, clinical reasoning~\cite{medxpertqa,slake}, and specialized domains~\cite{pathvqa,VQA-RAD,dermavqa}. For text-based QA, we use large-scale medical examinations~\cite{medQA,medmcqa}, MMLU clinical topics~\cite{mmlu}, and the text-only subset of MedXpertQA~\cite{medxpertqa} to evaluate the model’s advanced clinical reasoning abilities.

\textit{Evaluation Strategy.}
For general medical QA benchmarks, we follow the official evaluation protocols to ensure consistency and comparability with existing methods. For multiple-choice questions, a two-stage evaluation is adopted: exact rule-based matching first, and if no match is found, the answer with the highest semantic similarity is selected. For open-ended questions, GPT-4 is used for automated evaluation to judge the semantic consistency between the model’s response and the reference answer.

\subsection{Training Framework}
The training includes supervised fine-tuning on medical text and multi-modal data, followed by reinforcement learning tailored to diagnostic dialogue.
\subsubsection{Supervised Fine-Tuning.}
First, we train the model on Huatuo26M~\cite{huatuo26m}, to inject domain-specific knowledge and enhance its language understanding and clinical reasoning capabilities. Second, we further train the model on MediScope and public datasets to unlock its multimodal capabilities and multi-turn medical dialogue capabilities, thereby enhancing its ability to process clinical dialogues.

\subsubsection{Reinforcement Learning with CRPO.}
After SFT, the model acquires the ability to perform routine diagnostic consultations. To further enhance its human-aligned performance in real-world clinical consultation scenarios, we employ a RL approach to optimize the model across four key dimensions: proactiveness, accuracy, usefulness, and language quality. Popular RL methods like GRPO\cite{GRPO}, which rely on absolute scores as rewards, face inherent limitations in clinical dialogue scenarios. First, model-based absolute scores tend to be unstable and subjective. Second, whether model-based or rule-based, absolute scores often obscure differences among top-performing models, limiting their discriminative power.
To address these issues, we replace absolute score rewards with comparison-based relative rewards, and propose a Comparison-based Reinforcement Policy Optimization algorithm (CRPO) that facilitates a stable and human-aligned optimization process.

\begin{table}[ht]
\small
\setlength{\tabcolsep}{2.8pt}
\centering
\begin{tabular}{lccccc}
\toprule
Model & Proact. & Acc. & Use. & Lang. & Avg. \\
\midrule
Lingshu & 3.60 & 4.15 & 4.05 & 4.58 & 4.10 \\
Gemini2.5-pro & 3.91 & 4.35 & 4.30 & 4.69 & \underline{4.31} \\
o1    & 3.70 & 4.43 & 4.14 & 4.60 & 4.22 \\
GPT-4o    & 3.48 & 4.06 & 3.97 & 4.52 & 4.01 \\
Qwen2.5VL-72B & 3.81 & 4.30 & 4.22 & 4.59 & 4.23 \\
InternVL3-78B & 3.74 & 4.36 & 4.16 & 4.71 & 4.24 \\
PulseMind        & 3.92 & 4.47 & 4.26 & 4.763 & \textbf{4.35} \\
\bottomrule
\end{tabular}
\caption{Absolute scores assigned to seven models across four evaluation dimensions. The top two results are highlight with \textbf{bold} and \underline{underlined} fonts, respectively.}
\label{tab:abs}
\vspace{-1em}
\end{table}

Specifically, as illustrated in the lower-right part of Fig.~\ref{fig:framework}, given a sampled query \(q\), the policy model generates a set of candidate responses \(\{o_1, o_2, \ldots, o_G\}\). These responses are then evaluated by a reward model, which consists of a comparison model and five counterpart models \(\{CP_1, \ldots, CP_5\}\). The comparison model assesses each candidate response against those generated by the counterpart models across four key evaluation dimensions: proactiveness, accuracy, usefulness, and language quality.

For each candidate \(o_g\), the comparison model determines whether it performs better than each counterpart \(CP_c\) on each evaluation dimension \(d\). 
Concretely, the comparison model assigns a binary score:
\[
r_{g,c,d} =
\begin{cases}
1, & \text{if } o_g \succ CP_c \text{ on dimension } d, \\
0, & \text{otherwise},
\end{cases}
\]
where \(r_{g,c,d}\) indicates whether the \(g\)-th candidate response performs better than the \(c\)-th counterpart model on the \(d\)-th dimension.
The reward for the candidate \(o_g\) is obtained by averaging over all counterpart models and evaluation dimensions:
\[
R_g = \frac{1}{C \times D} \sum_{c=1}^C \sum_{d=1}^D r_{g,c,d},
\]
where \(C=5\) is the number of counterpart models and \(D=4\) is the number of evaluation dimensions. Following this, the advantage and final loss computation follow the same procedure as in GRPO.

\textit{Discussion.} 
To gain a clearer understanding of the differences between relative and absolute evaluation strategies, we conducted experiments examining their scoring behaviors and evaluation reliability. First, we analyze how absolute scoring performs in distinguishing different models. Second, we assess the reliability of the two strategies by comparing their consistency with human expert evaluations.

To evaluate the distribution and discriminative capacity of the absolute scoring strategy, we rated seven models using a 5-point scale across four evaluation dimensions: Proactiveness (Proact.), Accuracy (Acc.), Usefulness (Use.), and Language Quality (Lang.).
As shown in Tab.~\ref{tab:abs}, all models received relatively high and closely clustered scores, with average ratings ranging from 4.01 to 4.35. Such small score differences make it difficult to effectively differentiating the model performance.

To validate the effectiveness of the relative scoring approach, we randomly sampled 10\% of the evaluation outputs from both the absolute and relative scoring strategies and asked 50 medical experts to verify whether they agreed with each judgment. Based on their agreements, we computed the consistency rates between GPT-based evaluations and expert assessments across four dimensions as well as overall. As shown in Tab.~\ref{tab:cons}, the relative scoring strategy achieves an average consistency of 86.1\% on the PulseMind Benchmark, whereas the absolute scoring strategy reaches only 51.5\%, demonstrating the superior reliability of relative evaluation.

\begin{table}[t]
\centering
\footnotesize
\renewcommand{\arraystretch}{1.2}
\begin{tabular}{@{}l l|ccccc@{}}
\toprule
 &  & Proact. & Acc. & Use. & Lang. & All \\
\midrule
\multirow{2}{*}{Abs} 
& MedDi    & 53.7\% & 48.6\% & 48\% & 64.5\% & 42.1\%\\
& CMt   & 66.7\% & 58.1\% & 61.5\% & 73.4\% & 60.9\% \\
\midrule
\multirow{2}{*}{Rel} 
& MedDi     & 96\% & 88\% & 86.8\% & 92.2\% & 84.2\% \\
& CMt   & 98.4\% & 81.1\% & 94.1\% & 98.2\% & 87.9\%\\
\bottomrule
\end{tabular}
\caption{Consistency between GPT-based evaluations and human expert judgments under absolute (Abs) and relative (Rel) scoring strategies. MedDi and CMt denote the two subsets of the PulseMind Benchmark.}
\label{tab:cons}
\vspace{-1.0em}
\end{table}

\section{Experiments}
\subsection{Implementation Details}
We build on Qwen2.5-VL-72B and Qwen2.5-VL-32B. To achieve efficient parameter fine-tuning, we adopt a low-rank adaptation (LoRA) strategy by injecting a rank-64 adaptation matrix into the Transformer layer to freeze the base model while only training the newly added parameters.
Our training process is implemented on a cluster of 128 NVIDIA A100 GPUs and integrates a set of advanced optimization technology stacks. The technology stack is based on HuggingFace Transformers and PEFT libraries, combined with DeepSpeed ZeRO-3 to manage GPU memory, and BF16 mixed precision is enabled to accelerate calculations. The model is optimized using the AdamW optimizer, the learning rate is controlled by the cosine annealing scheduler, and a dropout rate of 0.1 is set to enhance generalization ability.

To evaluate performance, we benchmark against a wide range of models. General-purpose proprietary MLLMs include GPT-4o~\cite{gpt4}, o1~\cite{o1}, and Gemini 2.5 Pro~\cite{gemini}. Medical-specialized models include LLaVA-Med~\cite{llava-med}, HuatuoGPT-Vision~\cite{huatuo-vision}, and Lingshu~\cite{lingshu}. For broader comparison, we also include open-source general-purpose MLLMs such as InternVL3~\cite{internvl3} and Qwen2.5-VL~\cite{qwenvl}.

\begin{table*}[t]
\centering
\footnotesize
\renewcommand{\arraystretch}{1.2}
\begin{minipage}[t]{0.31\linewidth}
\centering 
\begin{tabular}{@{}lcc@{}}
\toprule
   & Public & Public+MediScope \\
\midrule
PulseMind-B  & 26.4  & 65.2           \\
MMMU      & 67.3 & 68.1           \\
VQA-RAD   & 86.6 & 86.9           \\
SLAKE     & 84.7 & 85.3           \\
MedXQA-MM & 34.9 & 36.5           \\
\bottomrule
\end{tabular}
\\ \textbf{(a) Dataset} 
\end{minipage}
\hspace{3em}
\hfill
\begin{minipage}[t]{0.31\linewidth}
\centering
\begin{tabular}{@{}lcc@{}}
\toprule
  & SFT  & SFT+RL         \\
\midrule
PulseMind-B  & 65.2 & 76.0           \\
MMMU      & 68.1 & 69.4          \\
VQA-RAD   & 86.9 & 87.1           \\
SLAKE     & 85.3 & 85.6           \\
MedXQA-MM & 36.5 & 36.6           \\
\bottomrule
\end{tabular}
\\ \textbf{(b) Training Stages} 
\end{minipage}
\hfill
\begin{minipage}[t]{0.31\linewidth}
\centering
\begin{tabular}{@{}lcc@{}}
\toprule
 & GRPO & CRPO           \\
\midrule
PulseMind-B  & 54.7   & 76.0          \\
MMMU      & 66.7   & 69.4           \\
VQA-RAD   & 86.9   & 87.1           \\
SLAKE     & 85.2   & 85.6           \\
MedXQA-MM & 36.1   & 36.6           \\
\bottomrule
\end{tabular}
\\ \textbf{(c) CRPO \textit{v.s.} GRPO} 
\end{minipage}
\caption{\textbf{PulseMind Ablation Experiments}. ``PulseMind-B'' denotes our PulseMind Evaluation Benchmark. ``Public'' indicates models trained on public datasets, while ``MediScope" refers to our dataset. ``SFT" and ``RL" denote using supervised fine-tuning or reinforcement learning, respectively. ``GRPO" and ``CRPO" represent different reinforcement learning strategies.}
\label{tab:ablations}
\end{table*}

\subsection{State-of-the-Art Comparisons}
The evaluation encompasses the PulseMind Benchmark as well as existing medical QA tasks, offering a comprehensive assessment of the models.
\subsubsection{PulseMind Benchmarks.} 
As shown in Fig.~\ref{fig:diagnose}, we evaluate the PulseMind model against leading models on the two subsets of the PulseMind Benchmark: MedDiagnose and cMtMedQA-test.

As shown in Fig.~\ref{fig:diagnose}(a), on the multi-modal MedDiagnose benchmark, our PulseMind model demonstrates superior performance.
Against proprietary general-purpose models, it achieves win rates of 94\% against GPT-4o, 89\% against o1, and 54\% against Gemini 2.5-Pro. For open-source general-purpose models such as Qwen2.5VL-72B and InternVL3, PulseMind attains win rates of 86\% and 83\%, respectively. When compared to the domain-specific medical model Lingshu, PulseMind shows a clear advantage with a win rate of 98\%.
These results underscore PulseMind’s capabilities in complex clinical scenarios.
As shown in Fig.~\ref{fig:diagnose}(b), on the expanded CMtMedQA-test benchmark, PulseMind maintains superior performance.
Against proprietary general-purpose models, it achieves win rates of 83\% with o1, 73\% with GPT-4o, and 72\% with Gemini 2.5-Pro.
For open-source general-purpose models, it secures 54\% against Qwen2.5VL-72B and 55\% against InternVL3.
When compared with the domain-specific medical model Lingshu, PulseMind attains a win rate of 71\%.
These results demonstrate PulseMind's robust generalization capability, even in text-only diagnostic tasks that emphasize broad medical knowledge.
In summary, PulseMind exhibits robust performance across both multi-modal and text-only multi-turn clinical consultation tasks, showcasing high adaptability to real-world clinical scenarios. Representative cases are shown in Fig.~\ref{fig:example} to illustrate model behaviors.

\subsubsection{Medical QA Benchmarks.}
As shown in Tab.~\ref{tab-multi}, PulseMind demonstrates leading performance on most medical QA benchmarks. Among open-source models with approximately 32B parameters, PulseMind-32B generally performs better than mainstream models such as Lingshu-32B and HuatuoGPT-vision-34B across most benchmarks. In comparisons involving larger-scale models, PulseMind-72B achieves the best results across all 11 benchmarks, surpassing peer open-source models and outperforming closed-source models in multiple tasks. 

\subsection{Ablation Study}
We conduct ablation studies using the PulseMind-72B model on the proposed PulseMind Benchmark as well as representative medical QA benchmarks to assess the impact of key design choices. The results are summarized in Tab.~\ref{tab:ablations}.

\textbf{Dataset.} As shown in Tab.~\ref{tab:ablations}(a), our curated multi-turn, multi-modality dataset, MediScope, brings substantial performance improvements on the PulseMind Benchmark. The average win rate (against six baseline models in Fig.~\ref{fig:diagnose}) increases significantly from 26.4\% to 65.2\%, highlighting the importance of constructing a heterogeneous, multi-source dataset grounded in realistic multi-turn diagnostic dialogues.
In addition, medical QA tasks also benefit from this dataset. For instance, the accuracy on MedXpertQA increases from 34.9\% to 36.5\%. Although the improvements on these public benchmarks are relatively modest, they demonstrate the strong generalization capability of MediScope beyond its primary target scenario.

\textbf{Training Stages.} As shown in Tab.~\ref{tab:ablations}(b), incorporating reinforcement learning further improves the model’s capability, raising the average win rate on PulseMind Benchmark from 65.2\% to 76.0\%. This highlights the effectiveness of reward-guided policy optimization in complex multi-turn dialogues. On medical QA benchmarks, reinforcement learning yields marginal gains, suggesting that supervised fine-tuning already sufficiently exploits the learning signal for these relatively simple tasks.

\textbf{CRPO \textit{v.s.} GRPO.} Table~\ref{tab:ablations}(c) compares our CRPO approach with the conventional GRPO strategy. CRPO demonstrates superior performance to GRPO on the PulseMind Benchmark, increasing the average win rate from 54.7\% to 76.0\%. It also shows improvements on other medical QA benchmarks. In particular, it achieves a 2.7\% improvement on the MMMU Health \& Medicine dataset.These results demonstrate the advantage of comparison-based relative rewards over absolute score-based rewards.

\section{Conclusion}
This work presents PulseMind, a multi-modal medical model for real-world clinical diagnosis. It includes a large-scale diagnostic dataset (MediScope), a diagnostic evaluation benchmark (PulseMind Benchmark), and a Comparison-based Reinforcement Policy Optimization method (CRPO). PulseMind demonstrates competitive performance on both the proposed benchmark and public medical benchmarks. We hope PulseMind can serve as a solid foundation for practical diagnostic dialogue applications.

\textit{Limitation.} PulseMind delivers strong multi-modal diagnostic capabilities, yet certain limitations persist. First, its ability to process specialized data formats, such as 3D medical imaging and other high-dimensional clinical modalities, remains limited. Second, training models demands substantial computational resources and considerable time, which may constrain their use in resource-limited environments.

\clearpage

\section{Acknowledgements}
This work is  supported by Ant Group Research Intern Program and National Natural Science Foundation of China under Grant 62431004.

\bibliography{aaai2026}

\end{document}